\documentclass[letterpaper, 10 pt, conference]{ieeeconf}  

\IEEEoverridecommandlockouts                              

\usepackage{algorithm}
\usepackage{algorithmic}
\usepackage{mathptmx} 
\usepackage{times} 
\usepackage{amsmath} 
\usepackage{amssymb}  
\usepackage{bm}
\usepackage{bbding}
\usepackage{xcolor}
\usepackage{array}
\newcolumntype{C}[1]{>{\centering\arraybackslash}p{#1}}
\usepackage{makecell}
\usepackage{hyperref}
\usepackage{url}
\usepackage{multirow}
\usepackage{graphicx}
\usepackage{subfig}
\usepackage{booktabs}

\newtheorem{definition}{Definition}

\newcommand{\kan}{KAN}
\newcommand{\model}{DMTP}
\newcommand{\data}{WOMD}

\title{\LARGE \bf
Scene-Aware Explainable Multimodal Trajectory Prediction
}

\author{Pei Liu, Haipeng Liu, Xingyu Liu, Yiqun Li, Junlan Chen, Yangfan He, and Jun Ma, \IEEEmembership{Senior Member, IEEE} 
\thanks{Pei Liu is with the Intelligent Transportation Thrust, The Hong Kong University of Science and Technology (Guangzhou), Guangzhou 511453, China (e-mail: pliu061@connect.hkust-gz.edu.cn).}%
\thanks{Haipeng Liu is with Li Auto Inc.,  Shanghai 201800, China (e-mail: liuhaipeng2012@live.com).}%
\thanks{Xingyu Liu is with the College of Land and Environment, Shenyang Agricultural University, Shenyang 110866, China (e-mail: 2022220462@test.syau.edu.cn).}%
\thanks{Yiqun Li and Junlan Chen are with the School of Transportation, Southeast University, Nanjing 211189, China (e-mail: 230248512@seu.edu.cn; junlan.chen@monash.edu).}
\thanks{Yangfan He is with the College of Libera Arts, University of Minnesota, Twin Cities, Minneapolis, MN 55455 USA (e-mail: he000577@umn.edu).}
\thanks{Jun Ma is with the Robotics and Autonomous Systems Thrust, The Hong Kong University of Science and Technology (Guangzhou), Guangzhou 511453, China, and also with the Division of Emerging Interdisciplinary Areas, The Hong Kong University of Science and Technology, Hong Kong SAR, China (e-mail: jun.ma@ust.hk). \textit{(Corresponding author: Jun Ma.)}  } 
      }

\begin{document}
\maketitle
\thispagestyle{empty}
\pagestyle{empty}
\begin{abstract}
Advancements in intelligent technologies have significantly improved navigation in complex traffic environments by enhancing environment perception and trajectory prediction for automated vehicles. However, current research often overlooks the joint reasoning of scenario agents and lacks explainability in trajectory prediction models, limiting their practical use in real-world situations. To address this, we introduce the Explainable Conditional Diffusion-based Multimodal Trajectory Prediction ({\model}) model, which is designed to elucidate the environmental factors influencing predictions and reveal the underlying mechanisms. Our model integrates a modified conditional diffusion approach to capture multimodal trajectory patterns and employs a revised Shapley Value model to assess the significance of global and scenario-specific features. Experiments using the Waymo Open Motion Dataset demonstrate that our explainable model excels in identifying critical inputs and significantly outperforms baseline models in accuracy. Moreover, the factors identified align with the human driving experience, underscoring the model's effectiveness in learning accurate predictions. Code is available in our open-source repository: \url{https://github.com/ocean-luna/Explainable-Prediction}.

\end{abstract}

\section{INTRODUCTION}
Motion prediction remains a critical challenge for autonomous vehicles, and it is crucial for safe navigation amidst uncertainties. In this context, motion prediction refers to forecasting the future trajectories of agents based on their past behaviors, contextual agents, road graphs, and traffic signals. Key challenges include the need for joint reasoning about the future trajectories of interacting agents, where naive independent predictions often yield unrealistic outcomes. While the past trajectory of the target agent is generally informative, interactions with other agents and traffic signs are often brief but significant. Deep learning advances, such as graph neural networks and transformers, have improved prediction capabilities, yet their black-box nature complicates understanding and interpreting the decision-making processes.

Traditional approaches in autonomous driving \cite{kato2018autoware, fan2018baidu} often treat trajectory prediction as a standalone module, limiting their ability to integrate global context and adapt to dynamic environments, and often lack interpretability. Justifying predictions through explanations is vital for understanding black-box models and building public trust \cite{omeiza2021explanations, atakishiyev2024explainable}. Prior efforts in using attention visualization for explaining deep learning models \cite{zablocki2021explainability, kim2017interpretable, hou2019interactive, kim2020attentional, zhou2021exploring} have mostly focused on single-modal predictions and intrinsic explainability.

To address these issues, we propose a trajectory prediction framework with improved explainability, called explainable conditional diffusion-based multimodal trajectory prediction. This model integrates a conditional diffusion approach to capture multi-agent interactions and an enhanced Shapley Value method to provide comprehensive global and scenario-specific explanations. Our contributions include:

\begin{itemize}
    \item Introducing the Explainable Conditional Diffusion-based Multimodal Trajectory Prediction ({\model}) framework, which enhances prediction accuracy while offering robust global and scene-based explainability.
\item Providing accurate predictions of multiple potential future positions for all agents within a scenario, incorporating detailed interaction information through the conditional diffusion model.
\item Delivering a comprehensive analysis on the value of explainability and the explanations for decisions.
Demonstrating alignment with human driving experience, which indicates the model’s robustness and effective learning of predictive behaviors.
\item Validating our approach with the extensive Waymo Open Motion Dataset ({\data}), showing superior performance compared to existing methods.
\end{itemize}

\section{RELATED WORK}
\subsection{Generative Models for Trajectory Prediction}
Recently, the growing need to accurately predict the trajectories of automated vehicles has spurred several advancements in generative modeling techniques. These approaches often focus on predicting motion by estimating conditional probabilities $p(s; c)$. For instance, HP-GAN \cite{barsoum2018hp} utilized an enhanced Wasserstein GAN to model the probability density function of future human poses based on previous poses.
Similarly, Conditional Variational Auto-Encoders (C-VAEs) \cite{oh2022cvae} and Normalizing Flows \cite{scholler2021flomo} have shown effectiveness in learning conditional probability density functions for predicting future trajectories. More recently, diffusion models have emerged as a promising alternative for modeling conditional distributions of sequences, including human motion poses \cite{zhang2024motiondiffuse} and planning tasks \cite{janner2022planning}. Notably, \cite{gu2022stochastic} leveraged diffusion models to capture uncertainties in pedestrian motion.

In diffusion-based models, multi-modal information is integrated using cross-attention mechanisms. For example, \cite{ngiam2021scene, amirloo2022latentformer} employed a cross-attention module where agent embeddings were used as queries, and map embeddings served as keys and values, facilitating interaction between map and agent tracks. Additionally, \cite{varadarajan2022multipath++} achieved state-of-the-art performance in the Argoverse Motion Forecasting Competition \cite{chang2019argoverse} through the use of multi-context gating (MCG) attention interaction modules. Furthermore, \cite{tang2022golfer} introduced a general diffusion-like architectural module, MnM, which secured second place in the 2022 {\data} Prediction Challenge \cite{schwall2020waymo}.

\subsection{Explainable Analysis Methods}

Deep learning systems have revolutionized numerous fields, but their opaque decision-making processes often present significant challenges. The lack of transparent explanations for their predictions can erode trust in these technologies. To address this, extensive research has been dedicated to developing explainable artificial intelligence methods \cite{jin2022enhancing, huang2023safari}, aimed at providing understandable explanations for model outputs.

In practical applications of explainable machine learning, two primary strategies are employed: using models that are intrinsically interpretable from the outset or applying post-hoc techniques to explain complex, pre-trained models. These strategies differentiate between intrinsic and post-hoc explainability \cite{moraffah2020causal}. Intrinsic interpretability involves models that are naturally understandable, such as decision trees or linear models. Conversely, post-hoc explainability refers to techniques developed to interpret the decisions of sophisticated models after training. This approach is further subdivided into global and local explainability. Global explainability seeks to reveal the general logic and internal mechanics of complex models, while local explainability focuses on elucidating the reasoning behind individual predictions for specific input instances.

\section{METHODOLOGY}
\begin{figure}
  \centering
  \includegraphics[width=0.5\textwidth]{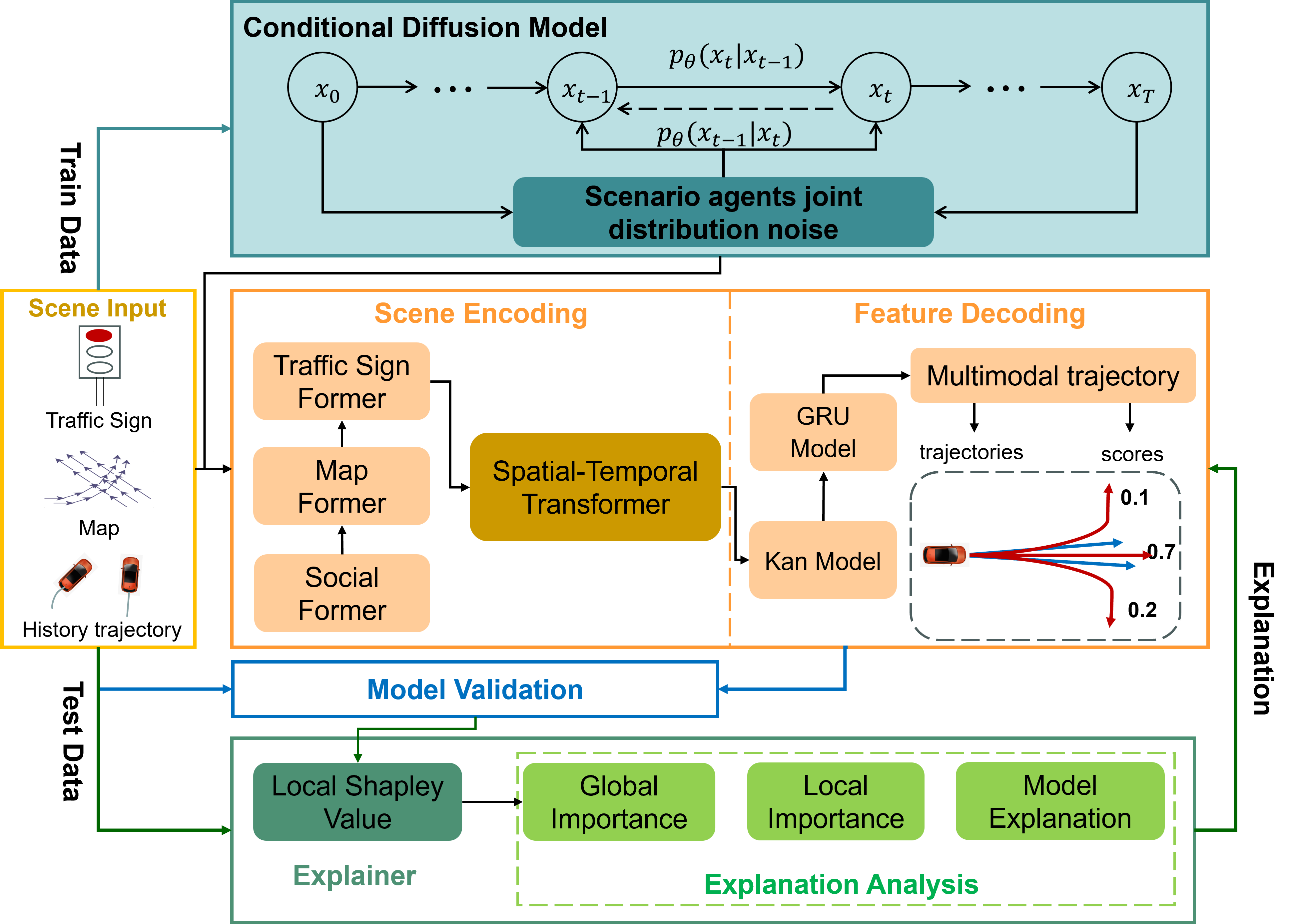} 
  \caption{Architecture of {\model}.}
  \label{fig:all}
\end{figure}

In this section, we present our innovative {\model} framework designed to model intricate traffic scenes while emphasizing feature importance. 
The framework is shown in Fig. \ref{fig:all}, including conditional diffusion, scene encoding,
feature decoding, and explanation analysis.

\subsection{Conditional Diffusion Model}
In traffic scenarios with multiple participants, using a single-agent diffusion model \cite{yang2024wcdt} can lead to suboptimal predictions due to the complex interactions and uncertainties among agents. To overcome this limitation, we propose a conditional diffusion model specifically designed to address these challenges. Our approach models trajectories as diffusion processes, which effectively capture the stochastic nature of movement patterns and adapt to dynamic environmental conditions and unexpected events. By incorporating trajectories of all agents in a scenario, our model accounts for spatial and temporal interactions among agents.

\begin{figure}[t]
  \centering
  \includegraphics[width=1\columnwidth]{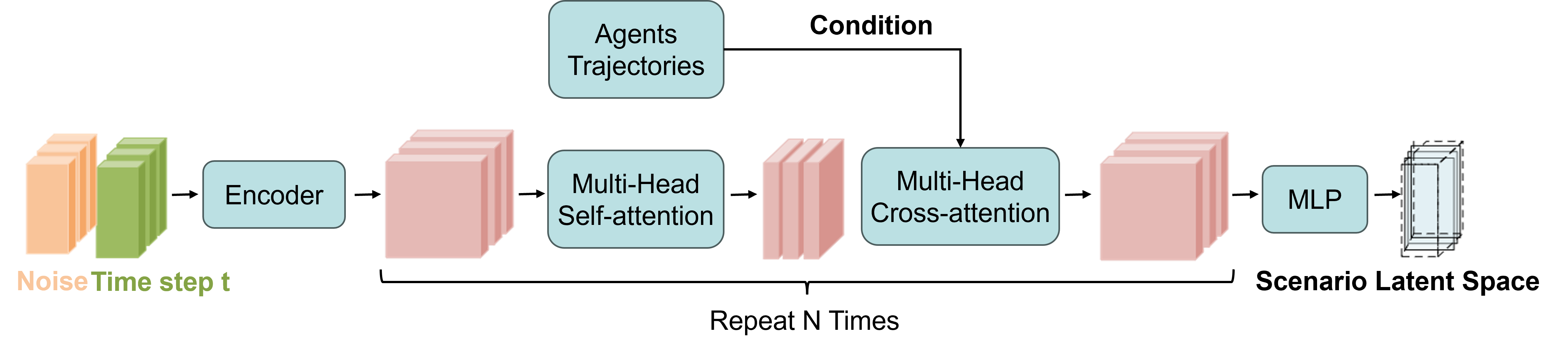} 
  \caption{Overview of the conditional diffusion for scenario latent space.}
  \label{fig:dits}
\end{figure}

Fig. \ref{fig:dits} shows the architecture of our scenario conditional diffusion blocks, which encode action latent information. The network takes random noise, temporal sequences, and agent trajectories as input, generating outputs in the scenario latent space, which are processed by the scene encoder. shows the architecture of our scenario conditional diffusion blocks, which encode action latent information. The network takes random noise, temporal sequences, and agent trajectories as input, generating outputs in the scenario latent space, which are processed by the scene encoder.
To enhance the performance of the diffusion network, we have meticulously designed the loss function for Denoising Diffusion Probabilistic Models (DDPM). This function guides the network to produce action latent representations that adhere to the kinematic constraints of agent behavior. This optimization ensures that the generated trajectories comply with physical laws and introduces variability to reflect the inherent uncertainty in the data produced by the diffusion model.
Our model samples trajectories at the scene level rather than the agent level, allowing it to simultaneously predict outcomes for all agents within a scene. This approach enables our model to effectively capture and account for interactions between agents, improving the accuracy of trajectory predictions in complex traffic scenarios. 

\subsection{Scene Encoding}
To effectively encode diverse agents, such as vehicles, bicycles, and pedestrians—along with map and traffic sign data, we employ multiple embedding blocks with varying sizes and layers. This design ensures the generation of diverse and realistic trajectories by capturing the comprehensive characteristics of each agent. Our approach eliminates the need for the traditional conversion of coordinates into Frenet coordinates centered around individual agents, a common practice in many existing trajectory prediction models.

In the encoder framework, each agent's positional and semantic embeddings are computed individually and processed through a multilayer perceptron to produce a unified embedding. This unified embedding is then enhanced with latent future embeddings generated by a diffusion model, resulting in predicted agent embeddings.
To create comprehensive scene-level representations that include predicted agents, other agents, map, and traffic sign, we apply cross-attention mechanisms. First, cross-attention is performed between the embeddings of the predicted agents and other agents to integrate information about agent interactions. Next, we apply cross-attention between these embeddings and map embeddings, sampled from various points within each map polygon, as well as traffic sign features represented as one-hot embeddings. These processes collectively ensure an accurate and detailed representation of the traffic scenes.

\textbf{Spatial and Temporal Fusion Attention.}
To effectively capture the complex dynamics of traffic scenes, we introduce a Temporal Spatial Fusion Attention (TSFA) layer. This layer is specifically designed to integrate temporal and spatial characteristics inherent in the movements of predicted agents, leveraging multi-modal data that includes predicted agents, neighboring agents, maps, and traffic signs within traffic environments.

To enhance the description of movements, the TSFA layer enriches its features with scenario latent features information derived from a diffusion process. Multi-head self-attention blocks are employed to extract critical insights from the spatial-temporal data. These blocks play a crucial role in identifying essential spatial and temporal details within the broader context of multi-modal traffic scenarios.

\subsection{Feature Decoding}
The trajectory decoder plays a crucial role in translating integrated traffic features into the future trajectories of agents. In this study, our decoder includes GRU blocks and KAN blocks, which are carefully configured to handle the temporal fluctuations inherent in agent movements. Inspired by the principles of multimodal trajectory prediction, which emphasize adaptability to agents exhibiting diverse behaviors over time, we introduce a multimodal output mechanism capable of effectively accommodating agents with varying actions in \cite{yang2024wcdt}.

\subsection{Explainability Analysis}
In this study, we aim to enhance our understanding of the input features used by trajectory generation models to achieve effective performance. Specifically, we focus on quantifying the impact of each input feature on the model's performance.

\subsubsection{Explainability of Model} 

From the perspective of information theory, higher model explainability means that we can obtain more useful information, which helps to improve the accuracy of the model.

\begin{definition}
    Information Gain is a measure of the amount by which a feature reduces the uncertainty of information. In information theory, information gain is obtained by calculating the entropy difference before and after the feature:
    \begin{equation}
        IG =H(Y) - H(Y|X)
    \end{equation}
where $H(Y)$ is the entropy of the target variable $Y$ (i.e., the overall uncertainty of the target variable). $H(Y|X)$ is the conditional entropy of the target variable $Y$ given feature $X$ (i.e., the reduction in uncertainty due to feature $X$).
\end{definition}
A highly explainable model can significantly reduce $H(Y|X)$, thereby increasing information gain and helping us better understand and optimize the model, thereby improving the accuracy of the model.

\subsubsection{Global and Scene Feature Importance}

\begin{definition}
\label{Markov}
\textbf{\textit{(Markov Blanket)}} \cite{koller1996toward} Given a feature $X_i$, the subset $M \subseteq \backslash X_i$ is a Markov blanket of $X_i$ if,
\begin{equation}
    p(\{F\backslash \{X_i, M\}, C\}|\{X_i, M\}) = p(\{F\backslash \{X_i, M\}, C\}|M)
\end{equation}
\end{definition}

This means that $M$ contains all the information about $C$ that $X_i$ has about $C$. It is proved that strongly relevant features do not have a Markov blanket. 

\begin{definition}
   \label{chain}
   \textbf{\textit{(Chain Rule for Mutual Information)}} \cite{cover1999elements}
    Given a set of random variables $X=\{X_1, X_2, \cdots, 
    X_n\}$ and random variable $Y$, then the mutual information of $X$ and $Y$ is defined as:
    \begin{equation}
        \begin{aligned}
            I(X,Y) & = I(X_1, X_2, \cdots, X_n;Y) \\ & =\sum_{y\in Y_{true}}\sum_{x\in X}p(x,y)\log \frac{p(x,y)}{p(x)p(y)} \\
            & = \sum_{i=1}^{n}I(X_i;Y|X_{i-1},X_{i-2},\cdots,X_1)
        \end{aligned}
    \end{equation}
\end{definition}
The chain rule for mutual information indicates the amount of information
that the random variables set $X$ can provide for $Y$ equals to the sum of pairwise
mutual information of $Y$ and each variable under certain conditions.

\begin{definition}
    \label{Relative}
     \textbf{\textit{(Relative Feature Importance)}} Let $X = (X_{1}, X_{2}, \cdots, X_{N})$ be the features of the to-be-predicted agent, and $Y$ is ground truth of the future trajectories of the to-be-predicted agent. We define relative feature importance of $X_i \in X$ with respect to $X \backslash X_i$ as:
    \begin{equation}
        RFI(X_j) = I(X_i;Y|X \backslash X_i)
    \end{equation}   
\end{definition}
$RFI(X_i)$ can be interpreted as the amount of reduced uncertainty of $Y$ due to $X_i$ given $X \backslash X_i$. $RFI(X_i) > 0$ means that $X_i$ is conditional relevance to 
$Y$. 

\begin{definition} 
\label{global}
\textbf{\textit{(Global Feature Importance)}}
    Let $X = (X_{1}, X_{2}, \cdots, X_{N})$ be the features of the to-be-predicted agent, and $Y$ is ground truth of the future trajectories of the to-be-predicted agent. From the perspective of information theory, the global feature importance of $X_i \in X$ is defined as:
\begin{equation}
    GFI(X_i) = I(X_i;Y) 
\end{equation}
\end{definition}
$GFI(X_i)$ can be interpreted as the amount of reduced uncertainty of $Y$ due to $X_i$. Global importance returns the overall impact of a feature on the model and is usually obtained by aggregating the feature attribution to the entire dataset. The higher the absolute value, the greater the impact of the feature on the model's predictions.

\begin{definition} 
\label{Scenario}
\textbf{\textit{(Scene Feature Importance)}}
    Let $x_{s} = (x_{1}, x_{2}, \cdots, x_{n})$ be the features local values of the to-be-predicted agent, and $y$ is ground truth of the future trajectories of the to-be-predicted agent in a specific scene. The Scene Feature Importance is defined as:
    \begin{equation}
        SFI = I(x_i,y|x_{s} \backslash x_i) 
    \end{equation}  
\end{definition}

Scene importance returns feature attribution values for each explained scene. These values describe how much a particular feature affects the prediction relative to the baseline prediction.

In our study, scene features are integrated with global features in the dataset, focusing on elements like historical trajectory, neighboring agents, the traffic sign (mainly traffic lights), and map, which align with the Markov blanket. The relationships between these features and the target are relative, as features rarely exist in isolation. To evaluate the model's representation capability, we introduce global and scene feature importance. Global importance is derived from high-order relative feature importance depicted by chain rule for mutual information, and we propose an approximate Shapley Value method for computing feature importance due to the impracticality of high-dimensional conditional probability calculations.

\subsection{Implementation of Feature Importance Measure}

\textbf{Scene Feature Importance Measure} For trajectory prediction, we modify the Shapley Value method by using a static, non-interacting agent as the baseline. This measures the impact of a target agent’s past trajectory relative to this baseline, with contributions from neighboring agents, traffic signs, and map data assessed in the absence of these elements.

\textbf{Global Feature Importance Measure}
Determining each feature’s overall contribution across varying scenes is challenging and inspired by \cite{makansi2021you}. We classify features into global and scene-specific types. Global features, such as past trajectories and map data, are evaluated using conventional averaging to estimate their importance. Scene-specific features, like neighboring agents and traffic signs, are aggregated in two steps: locally using the max operator within each scene to identify key elements, and globally using the average operator across the dataset to capture overall significance while accounting for variability.

\section{EXPERIMENTS}

\subsection{Evaluation Dataset}
We train and evaluate our model on a large-scale real-world dataset. {\data} \cite{ettinger2021large} contains 576,012 9-second sequences, each of which corresponds to a real trajectory sequence collected from the corresponding driving scene at a frequency of 10 Hz. Among them, (0,1] seconds are used for observation and (1,9] seconds are used for prediction.

\subsection{Evaluation Metrics}
We use the widely adopted official indicators.
For {\data}, we evaluate our model's performance using the Waymo Motion Prediction metrics. 

Following the standard evaluation protocol, we adopt The Minimum Average Displacement Error (minSADE), the Minimum Final Displacement Error (minSFDE), Missing Rate (sMR), and Mean average precision (mAP) for evaluation, which are defined in \cite{ettinger2021large}. 

These metrics serve as benchmarks to gauge the accuracy and reliability of our model's trajectory predictions against the {\data}.

\subsection{Quantitative Analysis}
To evaluate {\model}'s performance in multimodal trajectory prediction settings, we conduct our method against a range of leading-edge benchmarks on the {\data}. 

Table \ref{tab:comparison_results1} presents the primary metrics for multimodal trajectory prediction on the {\data}. We evaluate {\model} against various SOTA multimodal trajectory prediction methods, including SceneTransformer(M) \cite{ngiam2021scene}, SceneTransformer(J) \cite{ngiam2021scene}, Wayformer \cite{nayakanti2023wayformer}, MultiPath++ \cite{varadarajan2022multipath++}, JFP \cite{luo2023jfp}, MotionDiffuser \cite{ngiam2021scene}. While MotionDiffuser and our model are both diffusion-based approaches, our method outperforms significantly across all metrics, attributed to the robustness of our overall structure, with the exception of sMR. In comparison to JFP on both validation and test datasets, our model shows improvements in metrics such as minSADE, minSFDE, and mAP. 

\begin{table}[h]
    \centering
    \scalebox{0.85} {
    \begin{tabular}{cl|ccccc}
    \toprule[2pt]
        \multirow{9}*{\textbf{Val}} & \textbf{\makecell[c]{Method}}  & \textbf{\makecell[c]
        {minSADE} \ensuremath{\downarrow}} 
        & \textbf{\makecell[c]     
        {minSFDE}\ensuremath{\downarrow}} & \textbf{\makecell[c]        
        {sMR}\ensuremath{\downarrow}} & 
        \textbf{\makecell[c]
        {mAP}\ensuremath{\uparrow}}   \\
        \midrule[1pt]
        ~ & \textbf{\makecell[l]
        {SceneTransformer(M)}} & 1.12 & 2.60 & 0.54 & 0.09 \\
        ~ & \textbf{\makecell[l]
        {SceneTransformer(J)}} & 0.97 & 2.17 & 0.49 & 0.12 \\
        ~ & \textbf{Wayformer} &0.99 & 2.30 & 0.47 & 0.16 \\
        ~ & \textbf{MultiPath++ } & 1.00 & 2.33 & 0.54 & 0.18 \\
        ~ & \textbf{JFP} &  0.87 & 1.96 & \textbf{0.42} & 0.20 \\
        ~ & \textbf{MotionDiffuser} &  0.86 & 1.92 & \textbf{0.42} & 0.19 \\
        ~ & \textbf{\model(ours)} & \textbf{0.849} & \textbf{1.71} & 0.44 & \textbf{0.18}\\  
        \hline
        \multirow{4}*{\textbf{Test}} & \textbf{MultiPath++} & 1.00 &2.33 &0.54 &0.17\\
        ~ & \textbf{JFP} &0.88 &1.99 & \textbf{0.42} &0.21 \\
        ~ & \textbf{MotionDiffuser}  &0.86 &1.95 &0.43 &0.20\\
        ~ & \textbf{\model(ours)} &\textbf{0.85} & \textbf{1.74} & \textbf{0.42} & \textbf{0.20}\\
        \bottomrule[2pt]
    \end{tabular} }
    \caption{Evaluation of multimodal motion prediction capability on scene level joint metrics.}
    \label{tab:comparison_results1}
\end{table}

\subsection{Explainability Analysis}
This work examines the impact of historical trajectory, neighboring agents, the traffic sign, and the map features on the performance of the prediction algorithm. The minSADE and minSFDE are used as error metrics to evaluate model accuracy. The prediction model is trained on the {\data} training set, and an explainability analysis is conducted on its test set to better understand the influence of these features.

\subsubsection{Global Feature Importance}
\begin{figure}
    \begin{minipage}[t]{0.5\linewidth}
        \centering
        \includegraphics[width=1.0\textwidth]{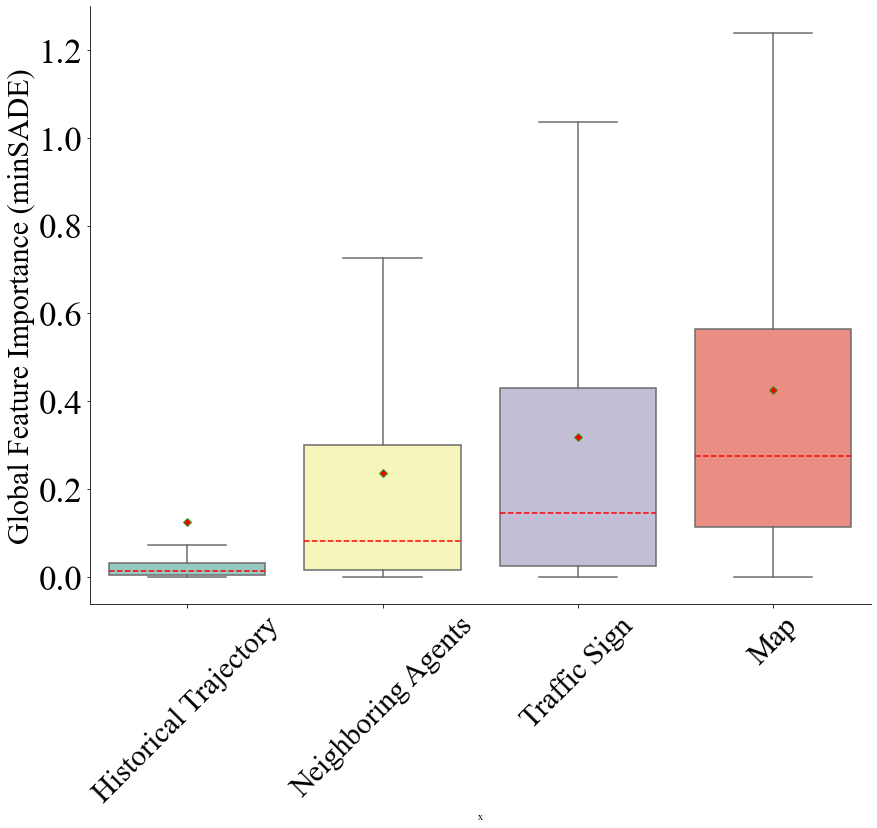}
    \end{minipage}%
    \begin{minipage}[t]{0.5\linewidth}
        \centering
        \includegraphics[width=1.0\textwidth]{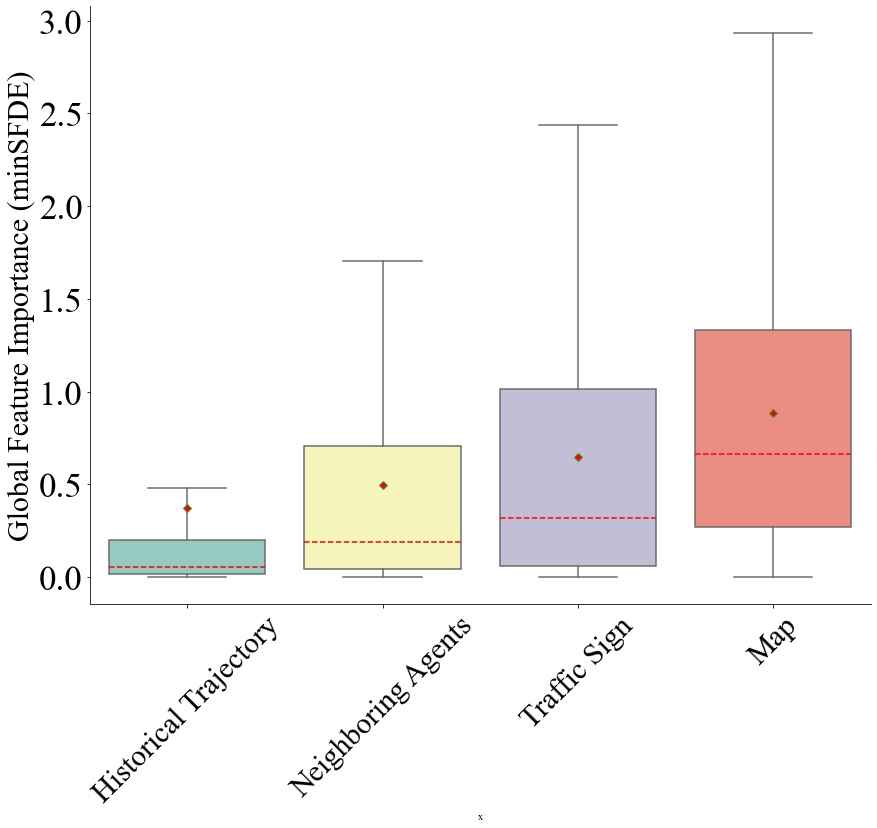}
    \end{minipage}
    \caption{Global feature importance (Shapley Value). The red dot represents the mean value and the red dashed line represents the median.} 
    \label{fig:global_importance}
\end{figure}

Fig. \ref{fig:global_importance} illustrates the feature importance for both minSADE error and minSFDE error. The results highlight that the map feature is the most influential, followed by the traffic sign, neighboring agents, and historical trajectory. This ranking aligns with human driving behavior cognition, suggesting that our model has effectively learned the correct semantic understanding of the features.
One might wonder why the historical trajectory holds the lowest importance. Inherently, this is a reasonable outcome, as the historical trajectory is a strongly correlated feature. However, when the map is provided, the trajectory becomes less informative and essentially becomes a redundant feature. This demonstrates that our model has successfully learned the redundant relationships between features, further validating its ability to handle feature dependencies effectively.

\subsubsection{Scene Feature Importance}
\begin{figure}[htbp]
    \begin{minipage}[t]{0.5\linewidth}
        \centering
        \includegraphics[width=1.0\textwidth]{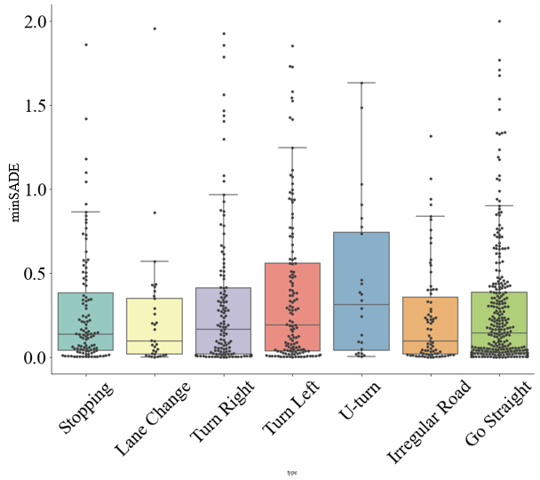}
    \end{minipage}%
    \begin{minipage}[t]{0.5\linewidth}
        \centering
        \includegraphics[width=1.0\textwidth]{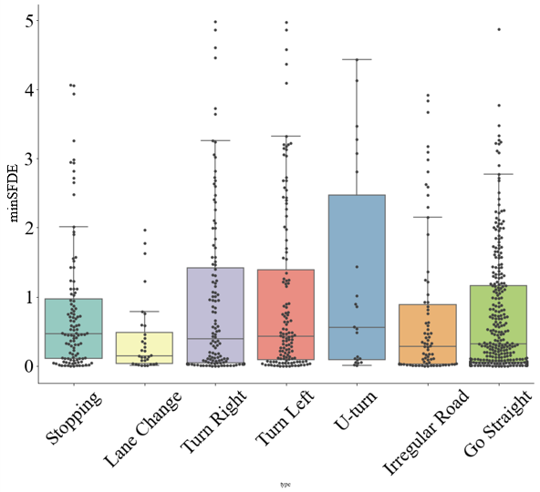}
    \end{minipage}
    \caption{The results of the trajectory prediction error under different traffic scenes caused by the traffic sign.} 
	\label{fig:Traffic}
\end{figure}

Fig. \ref{fig:Traffic} illustrates the varying importance of traffic signs across different driving scenarios. As shown, the traffic sign plays a critical role in scenarios such as stopping, turning left, and turning right, while their significance is minimal in irregular road conditions. This discrepancy can be attributed to the following factors: stopping, left turns, and right turns typically occur near intersections, where traffic signs directly influence decision-making. For instance, executing a right turn requires careful consideration of straight and left-turn signals, while stopping is influenced by opposing traffic signals and multiple crosswalks. In contrast, irregular roads often lack standardized traffic signs, resulting in their reduced importance in such scenarios. This analysis highlights the context-dependent nature of traffic sign relevance in autonomous driving systems.

\begin{figure}[!ht]
  \centering
  \includegraphics[width=1.0\linewidth]{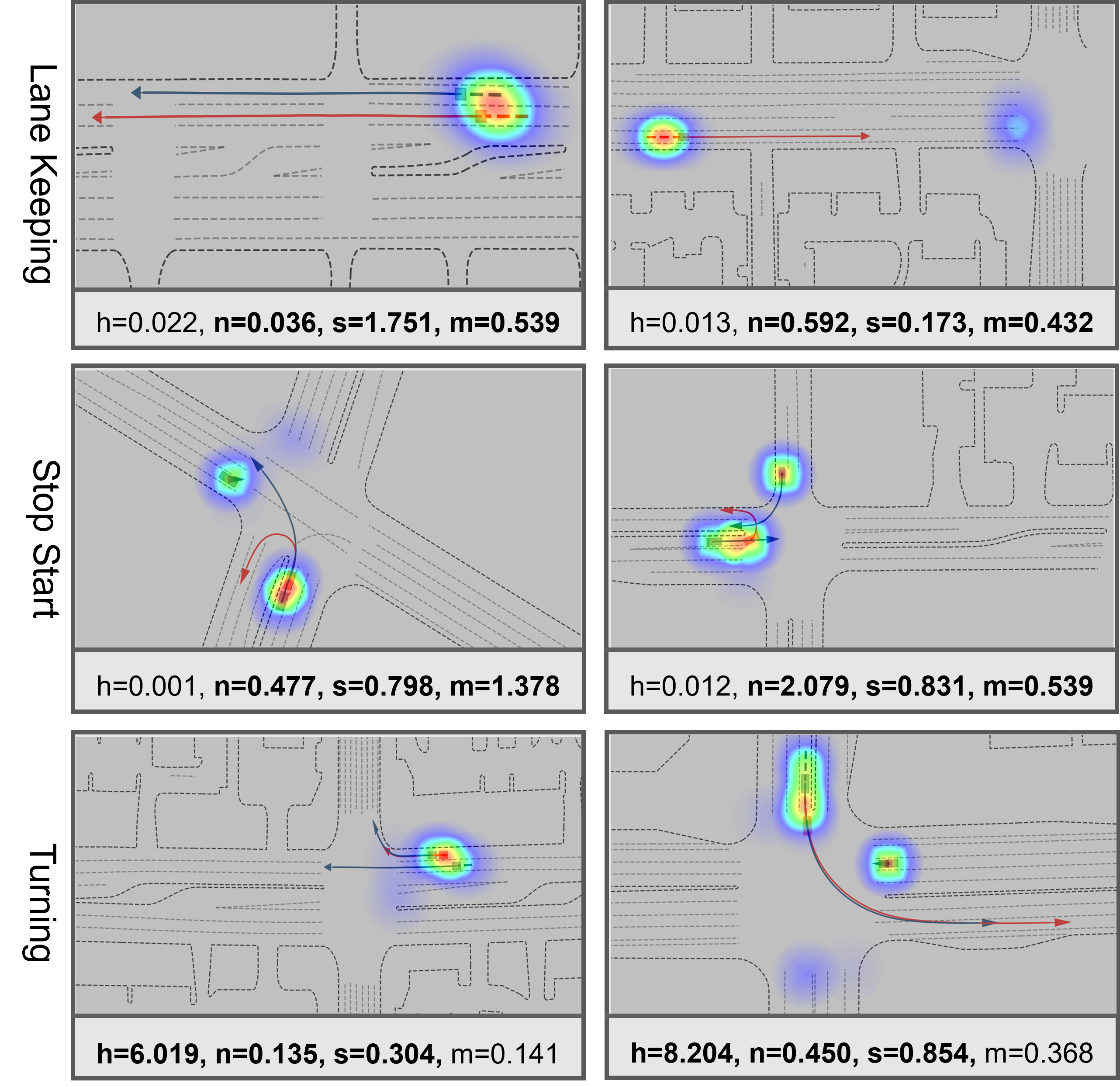} 
  \caption{Qualitative results of trajectory prediction and feature importance estimation on different scenarios (corresponding to different rows). Different colors are used to denote different agents: red - the predicted agent; blue - neighboring agents. The dotted line represents the historical trajectory, and the solid line represents the future trajectory. In addition, the scene feature importance is highlighted in the heatmap. h, n, s, m represent the feature importance of historical trajectory, neighboring agents, traffic sign, and map respectively.}
  \label{fig:local_importance}
\end{figure}

Fig. \ref{fig:local_importance} displays prediction outcomes across four scenarios: lane keeping, stop-start, turning, and interaction. In the first row, representing lane-keeping, predictions show high cognitive certainty, with future trajectories primarily driven by map data. The second and third rows highlight scenarios with significant prediction errors and higher cognitive uncertainty, especially at intersections or with major changes in movement. In stop-start scenarios, predictions depend more on neighboring agents, the traffic sign, and the map rather than the historical trajectory. Conversely, turning scenarios place a strong emphasis on historical trajectories, reflected by a high feature importance score of 8.204, aligning with the need for historical data due to discontinuous map lines at turns. The final row illustrates predictions at irregular intersections or road sections, characterized by the highest cognitive uncertainty, where future states are influenced mainly by neighboring agents and the map, with less reliance on traffic signs or historical data.

These results underscore conditional correlations between historical trajectory and map, highlighting instances where these features exhibit redundancy in predicting future trajectories. Moreover, our model effectively learns and reflects this redundancy in its scene feature importance rankings. For instance, in the first row of Fig. \ref{fig:local_importance}, either map features or historical trajectory exhibit lower importance depending on the scene. We also reached the same conclusion in terms of global feature importance. This confirms that our model is not only globally robust but also locally robust. 

\subsection{Ablation Studies on Traffic Scene Encoding}
In this subsection, we assess the structural impact by adding or removing components of the traffic scene encoder, which includes spatial and temporal attention modules, social former, map former, and traffic light former. We utilize minSADE and minSFDE metrics to evaluate the alignment of generated trajectories with ground truth.

Table \ref{table:ablation_encoder} presents the evaluation results, highlighting the critical role of the map former in accurate trajectory generation. Without this module, the model shows its poorest minSADE and minSFDE performances, indicating its crucial role in ensuring compliance with lane constraints during trajectory generation. Similarly, the spatial and temporal attention modules are essential, significantly enhancing the scene encoder's understanding of traffic scenes. Their absence results in minSADE and minSFDE scores of 1.122 and 1.993, respectively, below optimal levels.
The traffic sign module also proves crucial, as its absence negatively impacts trajectory accuracy to a degree comparable to missing spatial and temporal attention modules. Additionally, social former modules play a significant role in integrating the historical movements of neighboring agents, which is crucial for generating safe trajectories.

Combining all these modules, our traffic scene encoder achieves optimal performance, reflected in the lowest minSADE and minSFDE scores, demonstrating the effectiveness of our comprehensive approach in encoding the surrounding environment.

\subsection{Ablation Studies on Trajectory Decoding} 
In this subsection, we evaluate the architecture of the trajectory decoder, with a specific focus on GRU (Gated Recurrent Unit) blocks and {\kan} blocks. This evaluation aims to ascertain how each component and configuration impacts the decoder's ability to accurately predict future trajectories, emphasizing their critical role in enhancing model performance.

\begin{table}[t]
  \centering
  \scalebox{0.65}{
    \begin{tabular}{ccccc|cc}
      \toprule[2pt]
      \textbf{\makecell[c]{Spatial\&Temporal \\ Attention}} & 
      \textbf{\makecell[c]{Social Former}} & 
      \textbf{\makecell[c]{Map Former}} & 
      \textbf{\makecell[c]{Traffic Sign \\Former}} &  &
      \textbf{minSADE}\ensuremath{\downarrow} & \textbf{minSFDE}\ensuremath{\downarrow} \\
      \midrule[1pt]
        & \checkmark & \checkmark & \checkmark &  & 1.122 & 1.993\\
        \checkmark &            & \checkmark & \checkmark &  & 1.025 & 1.798\\
      \checkmark & \checkmark &            & \checkmark &  & 1.245 & 2.062\\
      \checkmark & \checkmark & \checkmark &            &  & 1.084 & 1.887\\
       
      \checkmark & \checkmark & \checkmark & \checkmark &  & \textbf{0.850} & \textbf{1.740}\\
      \bottomrule[2pt]
    \end{tabular}}
    \caption{Ablation study on the scene encoding.}
\label{table:ablation_encoder}
\end{table}

Table \ref{table:ablation-decoder} presents the evaluation results, highlighting the pivotal role of {\kan} layers in the trajectory decoder's performance. The absence of {\kan} blocks significantly deteriorates results, with the model achieving the highest minSADE and minSFDE scores of 1.990 and 2.857, respectively. This performance decline is attributed to the fundamental role of {\kan} blocks in improving the decoder's prediction capabilities and handling, integrating complex features extracted from input data.
Additionally, the importance of GRU blocks is underscored, enabling the decoder to effectively utilize historical information and further contribute to accurate future trajectory predictions. Combining {\kan} and GRU blocks, our model achieves optimal results, demonstrating outstanding trajectory prediction with the lowest minSADE and minSFDE scores.
Furthermore, we compared the performance of the {\kan} model and MLP model with experimental settings and results summarized in Table \ref{table:ablation-decoder}. It is evident that the Kan model with 2 layers achieves the best performance.

\begin{table}[h]
  \centering
  \scalebox{1.0}{
    \begin{tabular}{ccc|cc}
      \toprule[2pt]
      \textbf{\makecell[c]{MLP}} & \textbf{\makecell[c]{KAN}} & 
      \textbf{\makecell[c]{GRU}} &     
      \textbf{minSADE}\ensuremath{\downarrow} & 
      \textbf{minSFDE}\ensuremath{\downarrow} \\
      \midrule[1pt]
      &    & \checkmark & 1.990 & 2.857\\
      &  2 Layers &  & 0.892 & 2.170\\
     1 Layer &  & \checkmark & 0.861 & 1.803\\
    2 Layers  &  & \checkmark &  0.855 & 1.783\\
       &  1 Layer & \checkmark &  0.859 & 1.750\\
        & 2 Layers & \checkmark & \textbf{0.850} & \textbf{1.740}\\
      
      \bottomrule[2pt]
    \end{tabular}}
    \caption{Ablation study on the multimodal trajectory decoding.}
\label{table:ablation-decoder}
\end{table}

\section{CONCLUSION}
In this study, we introduce {\model}, a novel approach for Explainable Conditional Diffusion-based Multimodal Trajectory Prediction. Our method, {\model}, incorporates an advanced conditional diffusion technique to model diverse trajectory patterns. Additionally, we enhance the Shapley Value model to better understand global and scene-specific feature importance. Through extensive evaluations, we demonstrate the effectiveness of our approach on several benchmarks compared to other baselines. Future research directions involve extending the diffusion-based generative modeling technique to address other pertinent areas in autonomous vehicles, including planning conditional trajectory generation. We may also extend the framework for pedestrian trajectory prediction. 







\bibliographystyle{IEEEtran}
\bibliography{root}

\begin{thebibliography}{10}
\providecommand{\url}[1]{#1}
\csname url@samestyle\endcsname
\providecommand{\newblock}{\relax}
\providecommand{\bibinfo}[2]{#2}
\providecommand{\BIBentrySTDinterwordspacing}{\spaceskip=0pt\relax}
\providecommand{\BIBentryALTinterwordstretchfactor}{4}
\providecommand{\BIBentryALTinterwordspacing}{\spaceskip=\fontdimen2\font plus
\BIBentryALTinterwordstretchfactor\fontdimen3\font minus \fontdimen4\font\relax}
\providecommand{\BIBforeignlanguage}[2]{{%
\expandafter\ifx\csname l@#1\endcsname\relax
\typeout{** WARNING: IEEEtran.bst: No hyphenation pattern has been}%
\typeout{** loaded for the language `#1'. Using the pattern for}%
\typeout{** the default language instead.}%
\else
\language=\csname l@#1\endcsname
\fi
#2}}
\providecommand{\BIBdecl}{\relax}
\BIBdecl

\bibitem{kato2018autoware}
S.~Kato, S.~Tokunaga, Y.~Maruyama, S.~Maeda, M.~Hirabayashi, Y.~Kitsukawa, A.~Monrroy, T.~Ando, Y.~Fujii, and T.~Azumi, ``{Autoware on Board: Enabling Autonomous Vehicles with Embedded Systems},'' in \emph{2018 ACM/IEEE 9th International Conference on Cyber-Physical Systems}, 2018, pp. 287--296.

\bibitem{fan2018baidu}
H.~Fan, F.~Zhu, C.~Liu, L.~Zhang, L.~Zhuang, D.~Li, W.~Zhu, J.~Hu, H.~Li, and Q.~Kong, ``{Baidu Apollo EM Motion Planner},'' \emph{arXiv preprint arXiv:1807.08048}, 2018.

\bibitem{omeiza2021explanations}
D.~Omeiza, H.~Webb, M.~Jirotka, and L.~Kunze, ``{Explanations in Autonomous Driving: A Survey},'' \emph{IEEE Transactions on Intelligent Transportation Systems}, vol.~23, no.~8, pp. 10\,142--10\,162, 2021.

\bibitem{atakishiyev2024explainable}
S.~Atakishiyev, M.~Salameh, H.~Yao, and R.~Goebel, ``{Explainable Artificial Intelligence for Autonomous Driving: A Comprehensive Overview and Field Guide for Future Research Directions},'' \emph{IEEE Access}, 2024.

\bibitem{zablocki2021explainability}
{\'E}.~Zablocki, H.~Ben-Younes, P.~P{\'e}rez, and M.~Cord, ``{Explainability of Vision-based Autonomous Driving Systems: Review and Challenges},'' \emph{arXiv preprint arXiv:2101.05307}, vol.~2, 2021.

\bibitem{kim2017interpretable}
J.~Kim and J.~Canny, ``{Interpretable Learning for Self-Driving Cars by Visualizing Causal Attention},'' in \emph{Proceedings of the IEEE international conference on computer vision}, 2017, pp. 2942--2950.

\bibitem{hou2019interactive}
L.~Hou, L.~Xin, S.~E. Li, B.~Cheng, and W.~Wang, ``{Interactive Trajectory Prediction of Surrounding Road Users for Autonomous Driving Using Structural-LSTM Network},'' \emph{IEEE Transactions on Intelligent Transportation Systems}, vol.~21, no.~11, pp. 4615--4625, 2019.

\bibitem{kim2020attentional}
J.~Kim and M.~Bansal, ``{Attentional Bottleneck: Towards an Interpretable Deep Driving Network},'' in \emph{Proceedings of the IEEE/CVF Conference on Computer Vision and Pattern Recognition Workshops}, 2020, pp. 322--323.

\bibitem{zhou2021exploring}
J.~Zhou, R.~Wang, X.~Liu, Y.~Jiang, S.~Jiang, J.~Tao, J.~Miao, and S.~Song, ``{Exploring Imitation Learning for Autonomous Driving with Feedback Synthesizer and Differentiable Rasterization},'' in \emph{2021 IEEE/RSJ International Conference on Intelligent Robots and Systems}, 2021, pp. 1450--1457.

\bibitem{barsoum2018hp}
E.~Barsoum, J.~Kender, and Z.~Liu, ``{HP-GAN: Probabilistic 3D Human Motion Prediction via GAN},'' in \emph{Proceedings of the IEEE Conference on Computer Vision and Pattern Recognition Workshops}, 2018, pp. 1418--1427.

\bibitem{oh2022cvae}
G.~Oh and H.~Peng, ``{CVAE-H: Conditionalizing Variational Autoencoders via Hypernetworks and Trajectory Forecasting for Autonomous Driving},'' \emph{arXiv preprint arXiv:2201.09874}, 2022.

\bibitem{scholler2021flomo}
C.~Sch{\"o}ller and A.~Knoll, ``{FloMo: Tractable Motion Prediction with Normalizing Flows},'' in \emph{2021 IEEE/RSJ International Conference on Intelligent Robots and Systems}, 2021, pp. 7977--7984.

\bibitem{zhang2024motiondiffuse}
M.~Zhang, Z.~Cai, L.~Pan, F.~Hong, X.~Guo, L.~Yang, and Z.~Liu, ``{MotionDiffuse: Text-Driven Human Motion Generation With Diffusion Model},'' \emph{IEEE Transactions on Pattern Analysis and Machine Intelligence}, 2024.

\bibitem{janner2022planning}
M.~Janner, Y.~Du, J.~B. Tenenbaum, and S.~Levine, ``{Planning with Diffusion for Flexible Behavior Synthesis},'' \emph{arXiv preprint arXiv:2205.09991}, 2022.

\bibitem{gu2022stochastic}
T.~Gu, G.~Chen, J.~Li, C.~Lin, Y.~Rao, J.~Zhou, and J.~Lu, ``{Stochastic Trajectory Prediction via Motion Indeterminacy Diffusion},'' in \emph{Proceedings of the IEEE/CVF Conference on Computer Vision and Pattern Recognition}, 2022, pp. 17\,113--17\,122.

\bibitem{ngiam2021scene}
J.~Ngiam, B.~Caine, V.~Vasudevan, Z.~Zhang, H.-T.~L. Chiang, J.~Ling, R.~Roelofs, A.~Bewley, C.~Liu, A.~Venugopal \emph{et~al.}, ``{Scene Transformer: A Unified Architecture for Predicting Multiple Agent Trajectories},'' \emph{arXiv preprint arXiv:2106.08417}, 2021.

\bibitem{amirloo2022latentformer}
E.~Amirloo, A.~Rasouli, P.~Lakner, M.~Rohani, and J.~Luo, ``{LatentFormer: Multi-Agent Transformer-Based Interaction Modeling and Trajectory Prediction},'' \emph{arXiv preprint arXiv:2203.01880}, 2022.

\bibitem{varadarajan2022multipath++}
B.~Varadarajan, A.~Hefny, A.~Srivastava, K.~S. Refaat, N.~Nayakanti, A.~Cornman, K.~Chen, B.~Douillard, C.~P. Lam, D.~Anguelov \emph{et~al.}, ``{MultiPath++: Efficient Information Fusion and Trajectory Aggregation for Behavior Prediction},'' in \emph{2022 International Conference on Robotics and Automation}, 2022, pp. 7814--7821.

\bibitem{chang2019argoverse}
M.-F. Chang, J.~Lambert, P.~Sangkloy, J.~Singh, S.~Bak, A.~Hartnett, D.~Wang, P.~Carr, S.~Lucey, D.~Ramanan \emph{et~al.}, ``{Argoverse: 3D Tracking and Forecasting With Rich Maps},'' in \emph{Proceedings of the IEEE/CVF conference on computer vision and pattern recognition}, 2019, pp. 8748--8757.

\bibitem{tang2022golfer}
X.~Tang, S.~S. Eshkevari, H.~Chen, W.~Wu, W.~Qian, and X.~Wang, ``{Golfer: Trajectory Prediction with Masked Goal Conditioning MnM Network},'' \emph{arXiv preprint arXiv:2207.00738}, 2022.

\bibitem{schwall2020waymo}
M.~Schwall, T.~Daniel, T.~Victor, F.~Favaro, and H.~Hohnhold, ``{Waymo Public Road Safety Performance Data},'' \emph{arXiv preprint arXiv:2011.00038}, 2020.

\bibitem{jin2022enhancing}
G.~Jin, X.~Yi, W.~Huang, S.~Schewe, and X.~Huang, ``{Enhancing Adversarial Training With Second-Order Statistics of Weights},'' in \emph{Proceedings of the IEEE/CVF Conference on Computer Vision and Pattern Recognition}, 2022, pp. 15\,273--15\,283.

\bibitem{huang2023safari}
W.~Huang, X.~Zhao, G.~Jin, and X.~Huang, ``{SAFARI: Versatile and Efficient Evaluations for Robustness of Interpretability},'' in \emph{Proceedings of the IEEE/CVF International Conference on Computer Vision}, 2023, pp. 1988--1998.

\bibitem{moraffah2020causal}
R.~Moraffah, M.~Karami, R.~Guo, A.~Raglin, and H.~Liu, ``{Causal Interpretability for Machine Learning - Problems, Methods and Evaluation},'' \emph{ACM SIGKDD Explorations Newsletter}, vol.~22, no.~1, pp. 18--33, 2020.

\bibitem{yang2024wcdt}
C.~Yang, A.~X. Tian, D.~Chen, T.~Shi, and A.~Heydarian, ``{WcDT: World-centric Diffusion Transformer for Traffic Scene Generation},'' \emph{arXiv preprint arXiv:2404.02082}, 2024.

\bibitem{koller1996toward}
D.~Koller and M.~Sahami, ``{Toward Optimal Feature Selection},'' Stanford InfoLab, Tech. Rep., 1996.

\bibitem{cover1999elements}
T.~M. Cover, \emph{{Elements of Information Theory}}.\hskip 1em plus 0.5em minus 0.4em\relax John Wiley \& Sons, 1999.

\bibitem{makansi2021you}
O.~Makansi, J.~Von~K{\"u}gelgen, F.~Locatello, P.~Gehler, D.~Janzing, T.~Brox, and B.~Sch{\"o}lkopf, ``{You Mostly Walk Alone: Analyzing Feature Attribution in Trajectory Prediction},'' \emph{arXiv preprint arXiv:2110.05304}, 2021.

\bibitem{ettinger2021large}
S.~Ettinger, S.~Cheng, B.~Caine, C.~Liu, H.~Zhao, S.~Pradhan, Y.~Chai, B.~Sapp, C.~R. Qi, Y.~Zhou \emph{et~al.}, ``{Large Scale Interactive Motion Forecasting for Autonomous Driving: The Waymo Open Motion Dataset},'' in \emph{Proceedings of the IEEE/CVF International Conference on Computer Vision}, 2021, pp. 9710--9719.

\bibitem{nayakanti2023wayformer}
N.~Nayakanti, R.~Al-Rfou, A.~Zhou, K.~Goel, K.~S. Refaat, and B.~Sapp, ``{Wayformer: Motion Forecasting via Simple \& Efficient Attention Networks},'' in \emph{2023 IEEE International Conference on Robotics and Automation}, 2023, pp. 2980--2987.

\bibitem{luo2023jfp}
W.~Luo, C.~Park, A.~Cornman, B.~Sapp, and D.~Anguelov, ``{JFP: Joint Future Prediction with Interactive Multi-Agent Modeling for Autonomous Driving},'' in \emph{Conference on Robot Learning}, 2023, pp. 1457--1467.

\end{thebibliography}

\end{document}